\documentclass[]{xiaomiev}
\usepackage{iftex}
\ifxetex
\else
  \usepackage[utf8]{inputenc}
\fi

\usepackage{amsmath,amsfonts,amssymb}
\usepackage{nicefrac}

\usepackage{multicol}
\usepackage{colortbl}
\usepackage{makecell}
\usepackage{xltabular}
\usepackage{longtable}

\usepackage{tikz}
\usepackage{wrapfig}
\usepackage{adjustbox}
\usepackage{float}

\usepackage{enumitem}
\usepackage{xspace}
\usepackage{pifont}
\usepackage{listings}
\usepackage{fontawesome5}

\definecolor{abstractbar}{RGB}{210,140,50}
\definecolor{abstractbg}{RGB}{255,248,235}
\definecolor{BrickRed}{rgb}{.72,0,0}
\definecolor{darkgreen}{rgb}{0.0,0.5,0.0}
\definecolor{ForestGreen}{RGB}{34,139,34}
\definecolor{LakeBlue}{RGB}{0,61,153}
\definecolor{RoyalBlue}{RGB}{0,61,153}
\definecolor{MiOrange}{RGB}{255,225,204}
\definecolor{Hex}{RGB}{225,213,231}
\definecolor{xiaomiorange}{RGB}{255,103,0}
\definecolor{OliveGreen}{rgb}{0.33,0.42,0.18}

\newcommand{\METHODNAME}{{UI-MOPD}\xspace}
\newcommand{\BENCH}{{Uni-GUI}\xspace}

\newcommand{\METHODNAMEhl}{{\sffamily\bfseries\textcolor{xiaomiorange}{UI-MOPD}}\xspace}
\newcommand{\BENCHhl}{{\sffamily\bfseries\textcolor{RoyalBlue}{Uni-GUI}}\xspace}

\titleformat{\paragraph}[runin]
  {\normalfont\normalsize\bfseries\color[HTML]{FF7E00}}
  {}{0pt}{}
\titlespacing*{\paragraph}{0pt}{6pt}{1em}

\title{\METHODNAMEhl: Multi-Platform On-Policy Distillation for Unified GUI Agents}

\renewcommand{\authorfont}{\fontsize{10}{12}\selectfont}
\renewcommand{\affiliationfont}{\fontsize{8.5}{10}\selectfont}
\author[1,3]{Niu Lian\textsuperscript{*}}
\author[4]{Tongbo Chen\textsuperscript{*}}
\author[3]{Zhehao Yu}
\author[2]{Chengzhen Duan}
\author[2]{Fazhan Liu}
\author[2]{Hui Liu}
\author[2]{Pei Fu}
\author[2]{\\[-2pt]Jian Luan}
\author[2]{Heng Qu}
\author[1,5]{Shu-Tao Xia}
\author[3]{Jinpeng Wang\textsuperscript{\ding{72}}}
\contribution{\textsuperscript{*}Equal contribution.\quad \textsuperscript{\ding{72}}Corresponding author.\\[2pt] \href{mailto:220110904@stu.hit.edu.cn}{220110904@stu.hit.edu.cn} (Niu Lian), \href{mailto:wangjp26@gmail.com}{wangjp26@gmail.com} (Jinpeng Wang)}
\affiliation[1]{Tsinghua Shenzhen International Graduate School, Tsinghua University}
\affiliation[2]{Xiaomi\protect\\ $^{3}$Harbin Institute of Technology, Shenzhen}
\affiliation[4]{Zhejiang University}
\affiliation[5]{Peng Cheng Laboratory}

\resourcelinks{%
  \href{https://github.com/EliSpectre/UI-MOPD}{\faGithub~GitHub}%
  \hspace{1.5em}%
  \href{https://huggingface.co/EliSpctre/UI-MOPD}{\raisebox{-0.1em}{\includegraphics[height=0.95em]{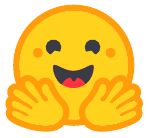}}~Hugging Face}%
  \hspace{1.5em}%
  \href{https://elispectre.github.io/UI-MOPD/}{\faGithub~Homepage}%
}

\renewcommand\checkdataformat[2][]{{\small \ifstrempty{#1}{}{{\checkdatafont\sffamily\bfseries #1:} }#2}}

\abstract{

Recent advances in multimodal foundation models and agent systems have
driven GUI agents from single-platform task execution toward cross-platform 
interaction. However, unified multi-platform GUI learning remains challenging: 
high-quality cross-platform trajectories remain scarce, while platforms share transferable capabilities but differ in action semantics and interaction conventions. Naively mixing supervision or merging specialized models
can blur native behaviors and produce imbalanced performance.
To address these challenges, we construct \BENCHhl, a
high-quality dataset containing nearly 10K executable cross-platform
interaction trajectories collected through a unified desktop-mobile
harness. Building on Uni-GUI, we propose \METHODNAMEhl, the first
framework to introduce multi-teacher on-policy distillation (MOPD) into unified
multi-platform GUI agent training.
UI-MOPD trains a shared student on its own rollouts and dynamically routes each rollout to the corresponding platform-specialized teacher. At student-visited states, teacher guidance serves as a platform-conditioned behavioral anchor, enabling the integration of complementary desktop and mobile expertise without averaging their distinct interaction conventions.
On OSWorld and MobileWorld, UI-MOPD achieves task success rates of 38.2\% and 12.0\%, respectively, outperforming parameter-matched integration strategies while preserving general GUI grounding. These results demonstrate that multi-teacher on-policy distillation provides an effective approach to building unified cross-platform GUI agents.

}

\begin{document}
\vspace*{-2cm}
\maketitle

\vspace{-11mm}
\begin{figure}[H]
  \centering
  \includegraphics[width=0.98\linewidth]{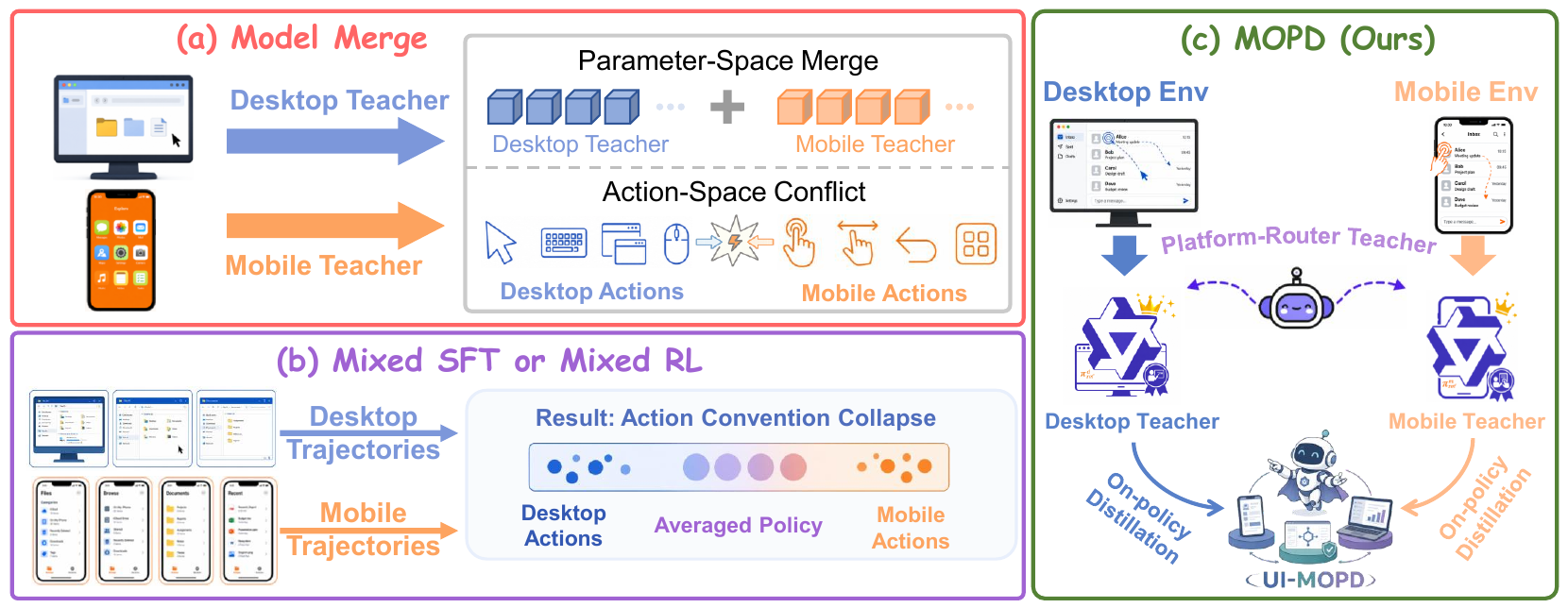}
  \caption{Motivation of UI-MOPD. Unlike model merging, mixed SFT, or mixed RL, which may yield an averaged policy, UI-MOPD integrates complementary platform expertise through multi-teacher on-policy distillation.}
  \label{fig:intro}
\end{figure}

\newpage
\tableofcontents
\newpage

\section{Introduction}
\label{sec: introduction}

Recent advances in multimodal foundation models~\citep{qwen3-vl,
Claude-Opus-4.6, Seed2.0} have strengthened visual understanding,
language reasoning, and tool-use capabilities~\citep{tu2026consensus}. Simultaneously, agent
systems~\citep{MM-MEM, milestone, Audio-Oscar} have evolved from
language only assistants toward interactive agents that can plan, invoke
tools, and operate in external environments. Graphical user interface
(GUI) agents~\citep{UI-TARS-2,Mobile-agent-v3.5} have therefore emerged
as a natural task form for connecting model intelligence with real digital
environments. GUI agents must understand screen content, plan operation
steps, and complete user goals through interface-level actions such as
clicking, typing, and swiping. Early studies largely centered on 
platform-specific web navigation or computer automation. Benchmarks such as OSWorld~\citep{osworld} and
MobileWorld~\citep{mobileworld} evaluate agents in interactive computer
and mobile environments, driving a shift from static interface
understanding toward long-horizon, platform-grounded interaction. Yet
real-world workflows often span computer applications, mobile apps, and web
services, requiring agents to adapt across heterogeneous GUI environments while preserving the interaction
conventions of each platform. This progression raises a central question:
\textbf{how can platform-specialized interaction knowledge be integrated into a unified GUI agent while preserving the behavioral conventions of each platform?}

Recent efforts extend GUI agents beyond isolated platforms by scaling interaction data or training shared policies across heterogeneous environments. Open-source datasets such as OpenCUA~\citep{Opencua} and OpenMobile~\citep{openmobile} provide large-scale GUI trajectories, while multi-platform GUI agents~\citep{Mobile-agent-v3.5, UI-venus-1.5} combine computer, mobile, and web signals through mixed supervised
fine-tuning (SFT), mixed reinforcement learning (RL), or model merging. However, these approaches largely view cross-platform learning as data aggregation or signal merging. We instead formulate it as a specialist knowledge integration problem, where platform-specific teachers provide complementary yet partially divergent corrections that simple data mixing, parameter averaging, or single-teacher training cannot effectively reconcile.

Desktop and mobile platforms share foundational capabilities such as visual understanding, GUI grounding, planning, and tool use, but differ in action semantics, affordances, and navigation conventions. For example, returning may mean closing a window on a computer but pressing the back button on mobile. Since these platforms are partly shared yet partly conflicting, neither isolated training nor indiscriminate signal averaging is ideal. Naively mixing heterogeneous supervision can blur platform-specific behaviors and induce capability conflation or imbalance. Cross-platform learning therefore requires balancing shared capability transfer with the preservation of platform-specific interaction patterns.

Our preliminary analysis reveals three key observations. First, desktop and mobile agents share transferable capabilities such as visual understanding, GUI grounding, and planning, yet conflict in action semantics and interaction conventions; consequently, specializing on a single platform risks degrading performance on the other. Second, naively integrating heterogeneous supervision fails to reconcile this conflict: both mixed supervised fine-tuning and static model merging tend to average away or overwrite platform-specific behaviors, degrading the weaker platform and general GUI grounding alike. Third, the expertise needed to resolve this conflict already resides in the platform-specific teachers; the key is therefore to apply each teacher as a behavioral anchor for its own platform, correcting the shared policy precisely at the states where that policy actually operates and tends to fail.

However, further validating these observations and enabling effective cross platform knowledge integration first requires high quality interaction data spanning multiple platforms, which remains scarce. Existing GUI datasets often focus on a single platform and may contain invalid actions, or inaccurate state action alignment. To address this limitation, we build a unified cross platform data collection harness for computer and mobile environments. Using this harness, we collect approximately $\mathbf{110K}$ computer and $\mathbf{50K}$ mobile interaction steps. After rigorous filtering, trajectory validation, and quality control, we construct \textbf{Uni-GUI}, a dataset containing nearly $\mathbf{10K}$ high quality and executable cross platform interaction trajectories.
Building on Uni-GUI, we propose \textbf{UI-MOPD}, the first method to 
introduce multi-teacher on-policy distillation (MOPD) into unified GUI agent learning, 
with platform-conditioned teachers for multi-platform adaptation.

UI-MOPD dynamically selects a platform-specific teacher for each
rollout environment and transfers platform-specific behavior distributions
to a shared policy. For computer and mobile environments, the model aligns
with native interaction patterns preserved by corresponding teachers,
preventing behavior signals with different interaction conventions from
being indiscriminately mixed during optimization. These platform-specific
teachers provide more than additional supervision. They serve as stable
behavioral anchors, allowing a shared policy to improve task completion
while preserving platform-specific behavioral priors. Through such
environment-conditioned policy alignment, UI-MOPD better balances
adaptation to new platforms with retention of existing platform behaviors,
mitigates behavioral convention mixing and the overwriting of
platform-specific behaviors, and encourages a
platform-aware policy that activates different interaction modes according
to environment context.

Empirical results further demonstrate the effectiveness of UI-MOPD. It achieves
task success rates of $\mathbf{38.2\%}$ on OSWorld and $\mathbf{12.0\%}$ on MobileWorld, corresponding to
relative improvements of $\mathbf{12.7\%}$ and $\mathbf{55.8\%}$ over the base model, respectively.
Importantly, these gains are observed in both computer and mobile environments,
suggesting that UI-MOPD improves adaptation to new platforms without sacrificing
existing platform capabilities and enables more balanced continual
optimization across platforms.

Main contributions of this work are summarized as follows:

\begin{itemize}
    \item We introduce \BENCHhl, a high-quality dataset for multi-platform GUI interaction. Using a unified cross-platform data-collection harness, we collect approximately 10K high-quality cross-platform interaction trajectories from computer and mobile environments.

    \item We propose \METHODNAMEhl, the first method to introduce multi-teacher on-policy distillation for unified GUI agent learning, addressing behavioral convention mixing and platform-specific degradation.

    \item We conduct systematic experiments on \textbf{OSWorld} and \textbf{MobileWorld}. UI-MOPD achieves task success rates of $\mathbf{38.2\%}$ and $\mathbf{12.0\%}$, respectively, demonstrating effective cross-platform capability preservation and adaptation.
\end{itemize}

\section{Related Work}
\label{sec:related}

\subsection{GUI Agent: From Single-plantform to Multi-plantform}

Graphical user interface (GUI) agents interpret natural language instructions and visual interface states to automate human interaction tasks in digital environments. Early GUI agent~\citep{xiaomigui0} research has largely focused on single-platform settings. Web environments provide online evaluation for browser-based navigation \citep{zhou2024webarena, koh2024visualwebarena}, where method such as WebVoyager improve web browsing. Mobile environments extend GUI agent evaluation to executable app control, cross-app workflows, personalization, and proactive assistance \citep{androidworld, mobileworld, Knowu-bench}, with methods such as UI-R1 \citep{UI-R1} and ClawGUI \citep{clawgui} exploring reinforcement learning. Desktop environments evaluate operating system level tasks \citep{osworld, bonatti2024windows, yang2026macosworld}, while recent computer-use agents such as EvoCUA \citep{evocua} and ComputerRL \citep{lai2025computerrl} study learning and decomposition for desktop automation. More recently, generalist GUI agents, including MobileAgent-v3.5~\citep{Mobile-agent-v3.5}, UI-Venus-1.5~\citep{UI-venus-1.5}, and UI-TARS-2~\citep{UI-TARS-2}, have explored unified perception, reasoning, and control across heterogeneous GUI platforms. However, directly combining data or training signals from different platforms may blur platform-specific interaction patterns and lead to capability interference. In contrast, \METHODNAME focuses on unified multi-platform GUI learning, aiming to train an end-to-end model for long-horizon navigation while effectively preserving, integrating, and transferring specialized interaction capabilities across desktop and mobile platforms.

\subsection{Multi-Teacher On-Policy Distillation}

During post training, different capabilities often exhibit a seesaw effect~\citep{scaling}: for example, mathematical RLVR may shorten reasoning traces and impair open ended writing, while each specialized training stage tends to improve one capability at the expense of others. To mitigate this issue, on policy distillation (OPD)~\citep{OPD, ExOPD, Uni-OPD} and on policy self distillation (OPSD)~\citep{OPSD, SDAR, SDFT} have emerged as effective solutions. OPD samples trajectories from the student model and matches the teacher distribution along these trajectories through reverse KL divergence, thereby providing dense token level supervision.
A natural extension of OPD is multi-teacher on-policy distillation (MOPD), which assigns the strongest checkpoint in each capability dimension as a teacher and enables the student model to absorb multiple capabilities in a single distillation stage. Recently, MOPD has been explored in foundation model post training. For example, MiMo-V2-Flash~\citep{Mimo-v2} and GLM-5~\citep{GLM-5} use MOPD as the final post training step to distill a unified model from multiple expert models. Nemotron-Cascade 2~\citep{Nemotron} uses MOPD as a forgetting recovery step between specialized RL stages, while DeepSeek-V4~\citep{DeepSeek-V4} further employs full vocabulary logits, more than ten teacher models, and dedicated infrastructure for teacher scheduling and fault tolerant trajectory generation.
However, MOPD remains largely unexplored in GUI agent. To our knowledge, \METHODNAME is the first to introduce MOPD into GUI
agent training, using platform-conditioned teachers to integrate desktop and
mobile interaction capabilities into a unified policy.

\section{Method}
\label{sec:method}

\subsection{Overview}
\label{sec:method_overview}

\METHODNAME trains a unified native graphical user interface (GUI) agent that can adapt to two 
heterogeneous environments, desktop and mobile, while preserving platform-specific interaction behaviors.
The construction of the unified cross-platform data collection harness and \BENCH is detailed in the Appendix.
The training procedure contains two stages. 
In Stage 1, we perform \textbf{supervised fine-tuning (SFT)} of a vision-language foundation model on high-quality trajectories 
from each platform, yielding two expert teachers: a desktop teacher $\pi_{\mathrm{ref}}^d$ and a mobile teacher $\pi_{\mathrm{ref}}^m$.
In Stage 2, we employ \textbf{multi-teacher on-policy distillation (MOPD)} to integrate the two specialized capabilities into a 
unified student model $\pi_\theta$.

\begin{figure}[h]
  \centering
  \includegraphics[width=\linewidth]{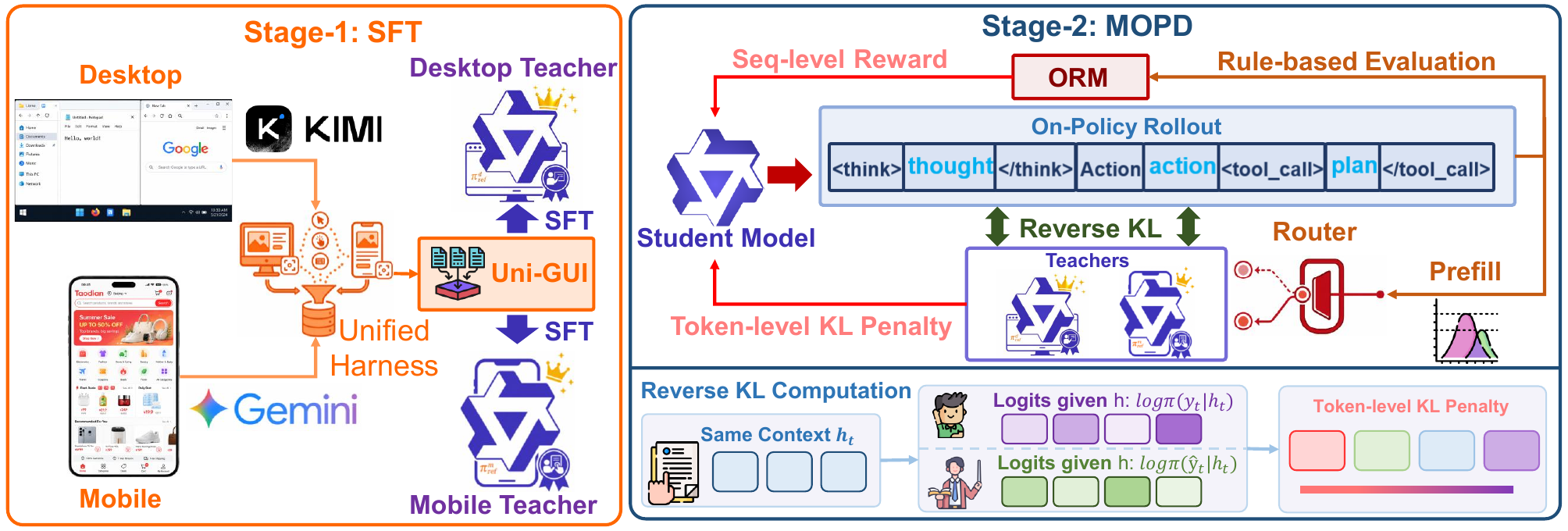}
  \caption{Overview of \METHODNAME training pipeline. In Stage 1, platform-specific desktop and mobile teachers are obtained by supervised fine-tuning on Uni-GUI trajectories collected from a unified cross-platform harness. In Stage 2, a shared student policy is trained with multi-teacher on-policy distillation, where platform-conditioned routing selects the corresponding teacher to provide reverse-KL guidance together with rule-based rollout rewards.}
  \label{fig:intro}
\end{figure}

\subsection{Multi-Teacher On-Policy Distillation}
\label{sec:mopd}

The core principle underlying \METHODNAME is to formulate multi-teacher knowledge
integration as a conditional behavioral constraint during online policy
optimization. Unlike directly merging multiple expert models or distilling
from static offline trajectories, we let the student policy $\pi_\theta$
sample rollouts online from its current policy and introduce teacher
supervision only on the states actually visited by the student. As a result,
the distillation signal is concentrated on the state distribution where the
current student policy truly makes decisions, rather than requiring the
student to imitate all behavioral modes of the teacher policies.
Specifically, teacher supervision is imposed in a platform-conditioned
manner: rollouts from different platforms are aligned with their
corresponding platform-specific teachers. In other words, teacher signals
from different platforms are not simply aggregated into a single
Kullback-Leibler (KL) penalty, but instead provide behavioral constraints
according to the platform to which each rollout belongs.

This design changes the role of the KL term from a conservative regularizer,
as commonly used in standard reinforcement learning from human feedback
(RLHF) to limit policy drift, into a more targeted mechanism for transferring
platform-specific expert interaction behaviors to a shared student policy
during online reinforcement learning. Since teacher supervision is applied to
the state distribution currently visited by the student, the resulting
distillation signal is better matched to the student policy's current errors.
This enables the shared policy to improve task success through reinforcement
learning while preserving the platform-specific behavioral anchors of both
desktop and mobile environments.

\paragraph{On-Policy Kullback--Leibler (KL).}
For the $i$-th sampled response $y^{(i)}=(y_1,\ldots,y_T)$, let
$h_t^{(i)}=(x^{(i)},y_{<t}^{(i)})$ denote the decoding state at token $t$.
Given the platform routed teacher $\pi_{\mathrm{ref}}^{(i)}$, we use the
student-to-teacher on-policy KL
\begin{equation}
D_{\mathrm{KL}}^{(t,i)}
=
D_{\mathrm{KL}}
\left(
\pi_\theta(\cdot \mid h_t^{(i)})
\;\|\;
\pi_{\mathrm{ref}}^{(i)}(\cdot \mid h_t^{(i)})
\right).
\label{eq:mopd_kl}
\end{equation}
Equivalently,
\begin{equation}
D_{\mathrm{KL}}^{(t,i)}
=
\mathbb{E}_{a \sim \pi_\theta(\cdot \mid h_t^{(i)})}
\left[
\log \pi_\theta(a \mid h_t^{(i)})
-
\log \pi_{\mathrm{ref}}^{(i)}(a \mid h_t^{(i)})
\right].
\label{eq:mopd_kl_expectation}
\end{equation}
The mini-batch MOPD loss is
\begin{equation}
\mathcal{L}_{\mathrm{MOPD}}(\theta)
=
\frac{
  \sum_{i \in B}\sum_t m_t^{(i)}\mu^{(i)}
  \hat{D}_{\mathrm{KL}}^{(t,i)}
}{
  \sum_{i \in B}\sum_t m_t^{(i)}\mu^{(i)}
},
\label{eq:mopd_loss}
\end{equation}
where $m_t^{(i)}$ masks out prompt and padding tokens, $\mu^{(i)}$ is an
adaptive KL mask, and $\hat{D}_{\mathrm{KL}}^{(t,i)}$ is a token level KL
estimate.

\paragraph{K3 Estimator.}
Computing the full KL over the vocabulary is expensive. We therefore use
the K3 estimator, which requires only the student and teacher log
probabilities of the sampled token. Define
\begin{equation}
\delta_t^{(i)}
=
\log \pi_{\mathrm{ref}}^{(i)}(y_t \mid h_t^{(i)})
-
\log \pi_\theta(y_t \mid h_t^{(i)}),
\qquad
\rho_t^{(i)}=\exp(\delta_t^{(i)}).
\label{eq:k3_delta}
\end{equation}
The token level estimate is
\begin{equation}
\hat{D}_{\mathrm{KL}}^{(t,i)}
=
\rho_t^{(i)}-\delta_t^{(i)}-1.
\label{eq:k3_estimator}
\end{equation}
This estimator is nonnegative, unbiased for
$D_{\mathrm{KL}}(\pi_\theta\|\pi_{\mathrm{ref}})$ under samples from
$\pi_\theta$, and empirically lower variance than direct log-ratio
estimators. We clamp $\delta_t^{(i)}$ in implementation for numerical
stability.

\paragraph{Adaptive KL Masking.}
Teacher constraints are most useful when task feedback is still weak. For
rollouts whose prompt group already receives sufficient reward, strong KL
regularization may unnecessarily restrict exploration. We therefore use a
group level mask
\begin{equation}
\mu^{(i)}
=
\begin{cases}
0, & \text{if } \frac{1}{G}\sum_{k\in g(i)} R^{(k)}
> \tau_{\mathrm{KL}},\\
1, & \text{otherwise},
\end{cases}
\label{eq:kl_mask}
\end{equation}
where $g(i)$ denotes the prompt group of sample $i$. This mask removes the
teacher penalty when task feedback is already sufficient for
policy improvement, while preserving teacher guidance on low reward
rollouts.

\subsection{Platform-Conditioned Teacher Routing}
\label{sec:teacher_routing}

Cross platform GUI interaction differs not only in visual layout, but also
in action semantics, affordance structure, and execution context. Desktop
tasks often require mouse operations, keyboard shortcuts, scrolling, and
window switching, whereas mobile tasks rely on tap, swipe, long press, and
application navigation. A single teacher or a direct mixture of teacher
logits would compress heterogeneous behaviors into an averaged
distribution, weakening platform specific interaction priors.

We instead train separate expert teachers and route each rollout by its
platform label:
\begin{equation}
\pi_{\mathrm{ref}}^{(i)}
=
\begin{cases}
\pi_{\mathrm{ref}}^{m},
& s_i \in \mathcal{S}_{\mathrm{mobile}},\\
\pi_{\mathrm{ref}}^{d},
& s_i \in \mathcal{S}_{\mathrm{desktop}},
\end{cases}
\label{eq:teacher_routing}
\end{equation}
where $s_i$ is the data source label recorded during data construction.
This routing affects only teacher log probability computation. The student
remains a single shared policy, so \METHODNAME does not introduce multiple
agents or additional teacher models at inference time.

During each reinforcement learning update, the student first samples
rollouts from mixed platform prompts. The mini-batch is then partitioned by
platform, each subset is evaluated by its corresponding teacher, and teacher
log probabilities are merged back into the original batch order. The K3
estimator in Eq.~\ref{eq:k3_estimator} then provides token level
distillation signals for Eq.~\ref{eq:mopd_loss}. This procedure gives each
platform a distinct behavioral anchor in the shared parameter space:
reinforcement learning moves the policy toward higher task reward, while
the routed teacher constrains this movement so that native interaction
patterns are not overwritten by signals from another platform.

\subsection{Reward Design}
\label{sec:reward_design}

We use a structured outcome reward for GUI actions. The policy outputs an
action JSON inside a tool-call format, including action type, coordinates,
text, or other action specific fields. For each action type, we define a set
of required dimensions, such as action type correctness, coordinate
inclusion in a target bounding box, scroll direction, key set equality, or
case insensitive text matching. Let $f_a\in[0,1]$ be the fraction of matched
dimensions for action $a$. The reward is
\begin{equation}
R(x,y)
=
\begin{cases}
1.0, & f_a = 1,\\
-0.5, & 0 \leq f_a < 1,\\
-1.0, & \text{unparsable or invalid action}.
\end{cases}
\label{eq:action_reward}
\end{equation}
The intermediate penalty distinguishes partially valid but incorrect
actions from invalid outputs, preserving a useful reward gap for
group-relative advantage estimation used in policy optimization.

\subsection{Training Objective}
\label{sec:training_objective}

Stage-2 training combines clipped policy optimization with the platform
conditioned MOPD penalty. Let
\begin{equation}
A_t^{(i)}
=
R(x^{(i)},y^{(i)})
-
\frac{1}{|g(i)|}\sum_{k\in g(i)}R(x^{(k)},y^{(k)})
\label{eq:advantage}
\end{equation}
be the token level advantage assigned to response tokens of sample $i$,
where $g(i)$ denotes its prompt group. This advantage is computed from the
structured reward in Eq.~\ref{eq:action_reward}; we write it as $A_t$ when
the sample index is clear. Let
\[
r_t(\theta)
=
\frac{
\pi_\theta(y_t\mid h_t)
}{
\pi_{\theta_{\mathrm{old}}}(y_t\mid h_t)
}
\]
be the policy ratio. For a rollout from platform $p$, we maximize the
regularized objective
\begin{equation}
\mathcal{J}(\theta)
=
\mathbb{E}_{p,x,y\sim\pi_\theta}
\left[
\sum_t m_t
\left(
\ell_{\mathrm{PG}}^{(t)}(\theta)
-
\beta\mu\,\hat{D}_{\mathrm{KL}}^{(t,p)}
\right)
\right],
\label{eq:regularized_objective}
\end{equation}
where
\begin{equation}
\ell_{\mathrm{PG}}^{(t)}(\theta)
=
\min
\left(
r_t(\theta)A_t,\,
\operatorname{clip}
\left(
r_t(\theta),
1-\epsilon_{\mathrm{low}},
1+\epsilon_{\mathrm{high}}
\right)A_t
\right).
\label{eq:policy_objective}
\end{equation}
Here $\hat{D}_{\mathrm{KL}}^{(t,p)}$ is computed with the teacher selected
by Eq.~\ref{eq:teacher_routing}, $m_t$ masks response tokens, $\mu$ is the
adaptive KL mask, and $\beta$ controls distillation strength. Equivalently,
we minimize
\begin{equation}
\mathcal{L}(\theta)
=
-
\mathcal{J}(\theta)
=
\mathcal{L}_{\mathrm{PG}}(\theta)
+
\beta\mathcal{L}_{\mathrm{MOPD}}(\theta).
\label{eq:final_loss}
\end{equation}
This objective improves long horizon task completion through online
reinforcement learning while preserving platform specific behavioral
anchors through routed teacher supervision.

\section{Experiments}
\label{sec:experiments}

\subsection{Overview}
\label{sec:exp_overview}

We evaluate whether \METHODNAME can train a single native GUI agent that
performs well across both desktop and mobile environments while preserving
general GUI understanding ability. The experiments are organized around
three questions:
\textbf{RQ1:} Do desktop and mobile specialists provide distinct and
complementary supervision?
\textbf{RQ2:} Where and when does specialist on-policy supervision provide
useful corrections to the student?
\textbf{RQ3:} Can routed multi-teacher OPD achieve balanced cross-platform
integration beyond single-teacher OPD and standard training strategies?

\subsection{Experimental Setup}
\label{sec:exp_setup}

\paragraph{Evaluation benchmarks.}
We evaluate cross-platform navigation and task execution on two interactive
benchmarks. OSWorld~\citep{osworld} contains 361 desktop tasks spanning
real computer applications, while MobileWorld~\citep{mobileworld} contains
117 tasks for evaluating mobile GUI interaction. We further assess static
GUI-agent capability on
AndroidControl\textsuperscript{$\star$}~\citep{androidcontrol}, where the
star denotes the evaluated subset whose construction is detailed in
Appendix~\ref{app:dataset_construction}. GUI grounding is evaluated on
ScreenSpot-Pro~\citep{Screenspot-pro}, ScreenSpotV2~\citep{screenspotv2},
and OSWorld-G~\citep{osworld-g}. We report task success rate for interactive
benchmarks and accuracy for static understanding and grounding benchmarks.

\paragraph{Evaluation protocols.}
We report per-platform success rates together with two balanced metrics.
The harmonic mean is
$\mathrm{H\mbox{-}Mean}=2S_{\mathrm{desktop}}S_{\mathrm{mobile}}/
(S_{\mathrm{desktop}}+S_{\mathrm{mobile}})$ and is omitted if either
platform is not evaluated. For controlled 8B comparisons, we additionally
report $\mathrm{MinGain}=\min(\Delta S_{\mathrm{desktop}},
\Delta S_{\mathrm{mobile}})$ relative to Qwen3-VL-8B-Thinking, measuring
the smaller improvement across the two platforms.

\paragraph{Training procedure and models.}
Training consists of two stages. In Stage 1, we supervised fine-tune
Qwen3-VL-32B-Thinking on the corresponding desktop and mobile partitions of
\BENCH to obtain two platform-specialized teachers. In Stage 2, the student
policy is initialized from Qwen3-VL-8B-Thinking and optimized with
reinforcement learning and MOPD. A single shared student is trained across
both environments, with each on-policy rollout routed to the teacher that
specializes in its platform. The two teachers are used only during training;
deployment requires only the resulting 8B student.

\paragraph{Implementation details.}
All training is implemented with verl~\citep{verl}, using
Megatron-Core~\citep{megatron} as the training backend and
SGLang~\citep{sglang} as the rollout engine. Experiments are conducted on
64 NVIDIA H100 GPUs organized as 8 nodes with 8 GPUs per node. During data
construction, desktop trajectories are collected with
Kimi-K2.6~\citep{kimi}, while mobile trajectories are collected with
Gemini-3.1-Pro~\citep{Gemini3.1}. Gemini-3.1-Pro additionally cleans both
sources under a unified filtering pipeline. For evaluation, all models are
deployed and served with vLLM~\citep{vllm}.

\subsection{Comparison with the State of the Art}
\label{sec:exp_main_results}

\begin{table}[!t]
\centering
\normalsize
\setlength{\tabcolsep}{3pt}
\renewcommand{\arraystretch}{0.90}
\setlength{\aboverulesep}{0.4ex}
\setlength{\belowrulesep}{0.45ex}
\begin{adjustbox}{max width=\linewidth}
\begin{tabular}{lccc}
\toprule
\raisebox{-0.40ex}{\textbf{Method}} & \raisebox{-0.40ex}{\textbf{OSWorld}} & \raisebox{-0.40ex}{\textbf{MobileWorld}} & \raisebox{-0.40ex}{\textbf{H-Mean}} \\
\midrule
\rowcolor{black!8}
\multicolumn{4}{l}{\textit{General Models}} \\
SeedVL-1.5 & 34.1\% & -- & -- \\
Qwen3-VL-4B-Instruct & 26.2\% & -- & -- \\
Qwen3-VL-4B-Thinking & 31.4\% & 6.8\% & 11.2\% \\
Qwen3-VL-8B-Instruct & 33.9\% & 9.4\% & 14.7\% \\
Qwen3-VL-8B-Thinking & 33.9\% & 7.7\% & 12.5\% \\
Qwen3-VL-32B-Instruct & 32.6\% & 9.0\% & 14.1\% \\
Qwen3-VL-32B-Thinking & 41.0\% & 9.4\% & 15.3\% \\
Qwen3-VL-235B-A22B-Instruct & 31.6\% & 9.5\% & 14.6\% \\
Qwen3-VL-235B-A22B-Thinking & 38.1\% & 12.0\% & 18.3\% \\
\midrule
\rowcolor{black!8}
\multicolumn{4}{l}{\textit{GUI Models (Single-Platform)}} \\
OpenCUA-7B & 28.2\% & -- & -- \\
OpenAI CUA o3 & 31.3\% & -- & -- \\
OpenCUA-32B & 34.8\% & -- & -- \\
\midrule
\rowcolor{black!8}
\multicolumn{4}{l}{\textit{GUI Models (Multi-Platform)}} \\
UI-TARS-72B-DPO & 27.1\% & -- & -- \\
UI-TARS-1.5-7B & 27.4\% & -- & -- \\
GELab-Zero-4B & 31.9\% & 10.9\% & 16.2\% \\
GUI-Owl-7B & 34.9\% & 4.5\% & 8.0\% \\
GUI-Owl-32B & -- & 5.5\% & -- \\
\midrule
\rowcolor{black!8}
\multicolumn{4}{l}{\textit{Controlled Training References}} \\
8B SFT on OSWorld & 35.8\% & 0\% & 0\% \\
8B SFT on MobileWorld & 35.8\% & 12.8\% & 18.9\% \\
Mixed-SFT (32B) & 44.4\% & 16.2\% & 23.7\% \\
Desktop Teacher (32B) & 46.3\% & -- & -- \\
Mobile Teacher (32B) & -- & 16.2\% & -- \\
\midrule
\rowcolor{black!8}
\multicolumn{4}{l}{\textit{Integration Strategies (8B Student)}} \\
Sequential-RL (Mobile $\rightarrow$ Desktop) & 36.6\% & 9.4\% & 15.0\% \\
Sequential-RL (Desktop $\rightarrow$ Mobile) & 33.7\% & 11.1\% & 16.7\% \\
Mixed-SFT (8B) & 35.0\% & 6.4\% & 10.8\% \\
Mixed-RL (8B) & 35.9\% & 9.4\% & 14.9\% \\
Unified-Teacher OPD (32B Mixed Teacher) & 35.9\% & 8.7\% & 14.0\% \\
Desktop-Only OPD (32B Teacher) & 36.0\% & 7.7\% & 12.7\% \\
Mobile-Only OPD (32B Teacher) & 34.9\% & 10.3\% & 15.9\% \\
MOPD (8B Teachers) & 36.2\% & 9.4\% & 14.9\% \\
Model Merge (Weight Averaging) & 36.5\% & 6.8\% & 11.5\% \\
Model Merge (TIES Merging) & 36.8\% & 0\% & 0\% \\
\midrule
\rowcolor{xiaomiorange!18}
\rule[-0.45ex]{0pt}{2.2ex}\raisebox{-0.20ex}{\textcolor{xiaomiorange}{\METHODNAME}} & \raisebox{-0.20ex}{\textbf{\textcolor{xiaomiorange}{38.2\%}}} & \raisebox{-0.20ex}{\textbf{\textcolor{xiaomiorange}{12.0\%}}} & \raisebox{-0.20ex}{\textbf{\textcolor{xiaomiorange}{18.3\%}}} \\
\bottomrule
\end{tabular}
\end{adjustbox}
\caption{Baselines, controlled training references, and integration
strategies on OSWorld and MobileWorld. Missing evaluations are marked as
``--''.}
\label{tab:main_results}
\end{table}

\begin{figure}[!t]
  \centering
  \captionsetup{font=small,skip=3pt}
  \includegraphics[width=0.92\linewidth]{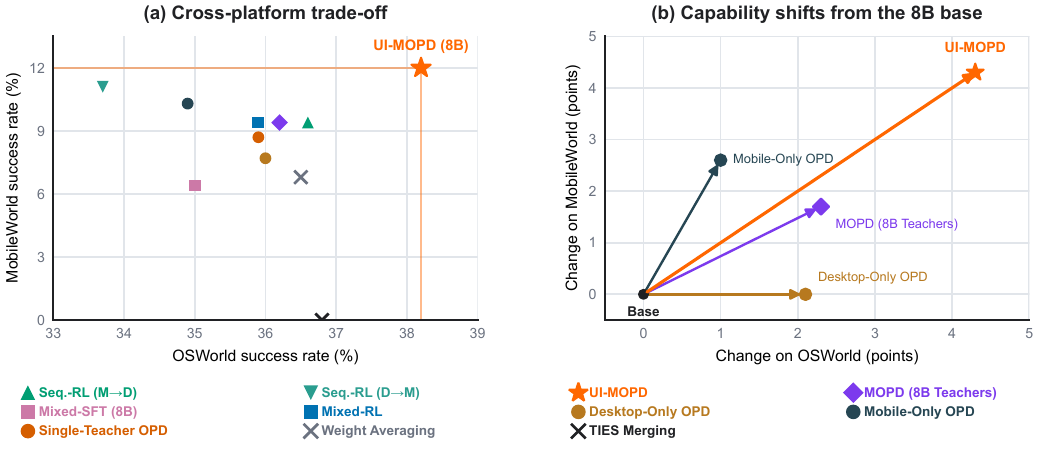}
  \caption{Cross-platform performance and capability shifts.
  \textbf{(a)} Trade-off among parameter-matched 8B integration strategies.
  \textbf{(b)} Gains over Qwen3-VL-8B-Thinking. Specialist teachers induce
  distinct shifts, while \METHODNAME yields the strongest joint improvement.}
  \label{fig:cross_platform_analysis}
\end{figure}

\paragraph{Baselines.}
Table~\ref{tab:main_results} compares \METHODNAME with representative
baselines in five groups. \textbf{General Models} include
SeedVL-1.5~\citep{seed1.5} and Qwen3-VL~\citep{qwen3-vl}. \textbf{GUI Models} cover
single-platform OpenCUA~\citep{Opencua} and OpenAI CUA~\citep{openai-cua},
and multi-platform UI-TARS~\citep{UI-TARS}, GELab-Zero~\citep{Step-gui},
and GUI-Owl~\citep{Mobile-agent-v3}. \textbf{Controlled Training
References} include single-platform 8B SFT, 32B Mixed-SFT, and two
platform-specialized 32B teachers. \textbf{Integration Strategies} train a
shared 8B policy via sequential or mixed SFT/RL, unified-teacher or
platform-only OPD, and MOPD with 8B teachers. The latter controls for
teacher scale, while weight averaging~\citep{Model-soups} and TIES
merging~\citep{Ties-merging} provide model-merging baselines.

\paragraph{Main analysis.}
Table~\ref{tab:main_results} and
Figure~\ref{fig:cross_platform_analysis} show that \METHODNAME achieves the best
balanced cross-platform performance among the compared 8B integration
strategies, reaching 38.2\% on OSWorld and 12.0\% on MobileWorld. Its
H-Mean of 18.3\% is the highest among the 8B integration strategies and
matches the much larger Qwen3-VL-235B-A22B-Thinking model. Compared with
general models, \METHODNAME improves MobileWorld performance substantially
over Qwen3-VL-8B-Thinking and Qwen3-VL-32B-Instruct, while also remaining
competitive on OSWorld. Existing GUI-agent baselines are often strong on one
side of the evaluation but incomplete or weaker on the other; for example,
GELab-Zero-4B obtains 10.9\% on MobileWorld but only 31.9\% on OSWorld,
whereas GUI-Owl-7B reaches 34.9\% on OSWorld but drops to 4.5\% on
MobileWorld. The parameter-matched 8B integration baselines further
indicate that directly mixing heterogeneous supervision, optimizing
platforms sequentially, relying on a single teacher, or statically merging
platform-specific models is insufficient for balanced integration at this
deployment scale. Sequential-RL exhibits an order-dependent trade-off,
Mixed-RL reaches 35.9\% and 9.4\%, and Unified-Teacher OPD reaches 35.9\%
and 8.7\% on OSWorld and MobileWorld, respectively. All parameter-matched
integration baselines underperform \METHODNAME on at least one platform.
These results suggest that platform-conditioned multi-teacher on-policy
distillation is more effective for transferring desktop and mobile expertise
into a single shared 8B GUI-agent policy.

\subsection{RQ1: Why Multiple Specialist Teachers?}
\label{sec:exp_specialists}

\begin{table}[t]
\centering
\small
\setlength{\tabcolsep}{3.2pt}
\renewcommand{\arraystretch}{0.92}
\begin{adjustbox}{max width=\linewidth}
\begin{tabular}{lcccc}
\toprule
\textbf{Method} & \textbf{OSWorld} & \textbf{MobileWorld} &
\textbf{H-Mean} & \textbf{MinGain} \\
\midrule
Mixed-RL (no teacher) & 35.9\% & 9.4\% & 14.9\% & 1.7 \\
Unified-Teacher OPD & 35.9\% & 8.7\% & 14.0\% & 1.0 \\
Desktop-Only OPD & 36.0\% & 7.7\% & 12.7\% & 0.0 \\
Mobile-Only OPD & 34.9\% & 10.3\% & 15.9\% & 1.0 \\
MOPD (8B Teachers) & 36.2\% & 9.4\% & 14.9\% & 1.7 \\
\METHODNAME w/o Adaptive Mask & 37.1\% & 10.3\% & 16.1\% & 2.6 \\
\METHODNAME w/o Specialist Routing & 36.4\% & 9.4\% & 14.9\% & 1.7 \\
\rowcolor{xiaomiorange!18}
\textcolor{xiaomiorange}{\METHODNAME} &
\textbf{\textcolor{xiaomiorange}{38.2\%}} &
\textbf{\textcolor{xiaomiorange}{12.0\%}} &
\textbf{\textcolor{xiaomiorange}{18.3\%}} &
\textbf{\textcolor{xiaomiorange}{4.3}} \\
\bottomrule
\end{tabular}
\end{adjustbox}
\caption{Ablation studies of teacher specialization, teacher scale,
platform-conditioned routing, and adaptive KL mask.}
\label{tab:specialist_ablation}
\end{table}

\paragraph{Ablation setup.}
Table~\ref{tab:specialist_ablation} isolates teacher specialization,
teacher routing, and adaptive KL masking while keeping the 8B student and
Stage-2 training data fixed. Desktop-Only OPD moves the student mainly along
the OSWorld axis, whereas Mobile-Only OPD produces a larger shift on
MobileWorld (Figure~\ref{fig:cross_platform_analysis}(b)). Neither specialist
alone matches the joint improvement of \METHODNAME. More importantly,
Unified-Teacher OPD uses a strong 32B teacher trained on mixed
desktop--mobile data, yet reaches only 35.9\%/8.7\%. Thus, the advantage of
\METHODNAME cannot be explained by ordinary OPD or teacher scale alone.
Replacing the 32B specialists with 8B teachers also weakens the result,
showing that both specialization and teacher quality matter.

\paragraph{Teacher routing and adaptive masking.}
Randomly selecting between the two teachers removes the platform--expert
correspondence and substantially reduces the balanced gain. Applying KL to
all rollout groups also underperforms the full method: once a group already
receives sufficient task reward, continued imitation can unnecessarily
restrict policy improvement. Together, these controls support routed
specialists rather than an undifferentiated teacher constraint.

\subsection{RQ2: Where Does Specialist Supervision Help?}
\label{sec:exp_corrections}

Final success rates do not reveal what information is transferred from each
teacher. We therefore analyze teacher--student action disagreement on states
visited by the current student. A disagreement is counted as a
\emph{useful correction} when the teacher proposes a valid target action
while the sampled student action is incorrect; it is \emph{harmful} when
the reverse holds. Equivalent executable actions are treated as agreement.

\begin{figure}[t]
  \centering
  \captionsetup{font=small,skip=3pt}
  \includegraphics[width=0.96\linewidth]{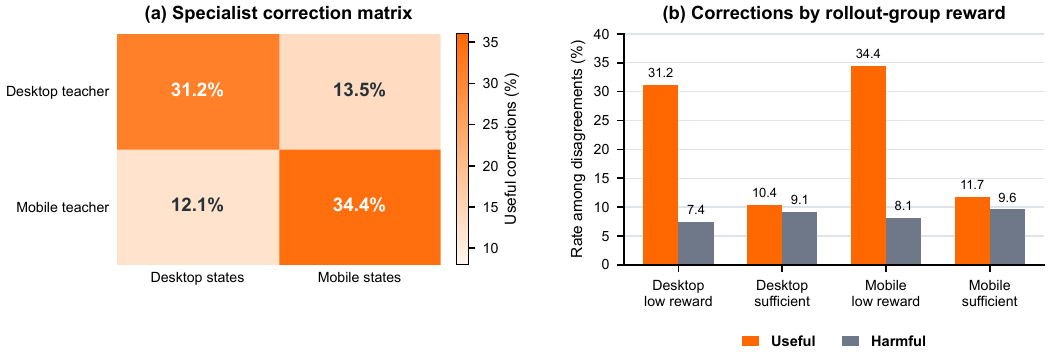}
  \caption{Teacher corrections at student-visited states.
  \textbf{(a)} Useful correction rate conditioned on action disagreement;
  each specialist is most effective on its native platform.
  \textbf{(b)} Routed teachers provide more useful than harmful corrections
  in low-reward rollout groups, whereas the gap narrows after task feedback
  becomes sufficient.}
  \label{fig:teacher_corrections}
\end{figure}

Figure~\ref{fig:teacher_corrections}(a) exhibits a diagonal pattern: the
desktop teacher supplies more useful corrections on desktop states, while
the mobile teacher is more effective on mobile states. This provides
state-level evidence that the two teachers contain complementary behavioral
knowledge rather than duplicated supervision.
Figure~\ref{fig:teacher_corrections}(b) shows that useful disagreement is
concentrated in low-reward rollout groups, precisely where the adaptive mask
retains teacher guidance. On successful groups, the useful--harmful gap
becomes much smaller, supporting the selective use of OPD rather than
uniform imitation.

\subsection{RQ3: Balanced Integration and Retention}
\label{sec:exp_balanced}

\METHODNAME gains 4.3 points on both OSWorld and MobileWorld, giving the
largest MinGain in Table~\ref{tab:specialist_ablation}. By contrast,
sequential RL depends on training order, single-platform OPD favors its
native platform, and the unified teacher provides only limited gains.

\paragraph{Cross-platform retention.}
The controlled references in Table~\ref{tab:main_results} further show why
improving both platforms is non-trivial. Desktop-only SFT raises the 8B base
model from 33.9\% to 35.8\% on OSWorld but collapses MobileWorld performance
from 7.7\% to 0\%. Mobile-only SFT reaches 12.8\% on MobileWorld, but its
35.8\% OSWorld score remains below the 38.2\% achieved by \METHODNAME.
In contrast, \METHODNAME raises both base scores by 4.3 points and even
surpasses the Qwen3-VL-32B-Thinking base model on MobileWorld. The 32B
Mixed-SFT result is a capacity reference rather than a parameter-matched
baseline: it shows that a substantially larger model can absorb mixed
supervision, whereas \METHODNAME targets a single 8B policy at deployment.

\paragraph{General GUI retention.}
To assess whether these interactive gains preserve broader GUI capability,
we evaluate static understanding and grounding after cross-platform policy
optimization.
Table~\ref{tab:grounding} reports results on
AndroidControl\textsuperscript{$\star$}, ScreenSpot-Pro, ScreenSpotV2, and
OSWorld-G.

\begin{table}[t]
\centering
\normalsize
\begin{adjustbox}{max width=\linewidth}
\begin{tabular}{lcccc}
\toprule
\textbf{Model} & \textbf{AndroidControl\textsuperscript{$\star$}} &
\textbf{ScreenSpot-Pro} & \textbf{ScreenSpotV2} & \textbf{OSWorld-G} \\
\midrule
Qwen3-VL-8B-Thinking & 78.73\% & 43.71\% & 91.27\% & 52.13\% \\
Mixed-SFT (8B) & 75.86\% & 39.92\% & 89.43\% & 48.76\% \\
Mixed-RL (8B) & 78.11\% & 42.58\% & 90.69\% & 51.21\% \\
Unified-Teacher OPD & 78.64\% & 42.91\% & 90.94\% & 51.77\% \\
Model Merge (TIES Merging) & 74.01\% & 37.13\% & 88.60\% & 47.16\% \\
\METHODNAME & \textbf{80.05\%} & 43.14\% & 90.88\% & \textbf{52.84\%} \\
\bottomrule
\end{tabular}
\end{adjustbox}
\caption{General GUI grounding, visual understanding, and
AndroidControl\textsuperscript{$\star$} results. The star denotes the
evaluated AndroidControl subset; Model Merge uses TIES merging.}
\label{tab:grounding}
\end{table}

\METHODNAME improves AndroidControl\textsuperscript{$\star$} from 78.73\%
to 80.05\% and OSWorld-G from 52.13\% to 52.84\%, while remaining within
0.6 points of the base model on ScreenSpot-Pro and ScreenSpotV2. In
contrast, mixed SFT and parameter merging cause broader degradation, and
ordinary RL or unified-teacher OPD does not match \METHODNAME's interactive
gains. These results indicate that routed on-policy supervision integrates
platform behaviors without broadly overwriting static GUI grounding.

\subsection{Mobile Case Study}
\label{sec:exp_case_study}

\begin{figure}[H]
  \centering
  \includegraphics[height=0.80\textheight]{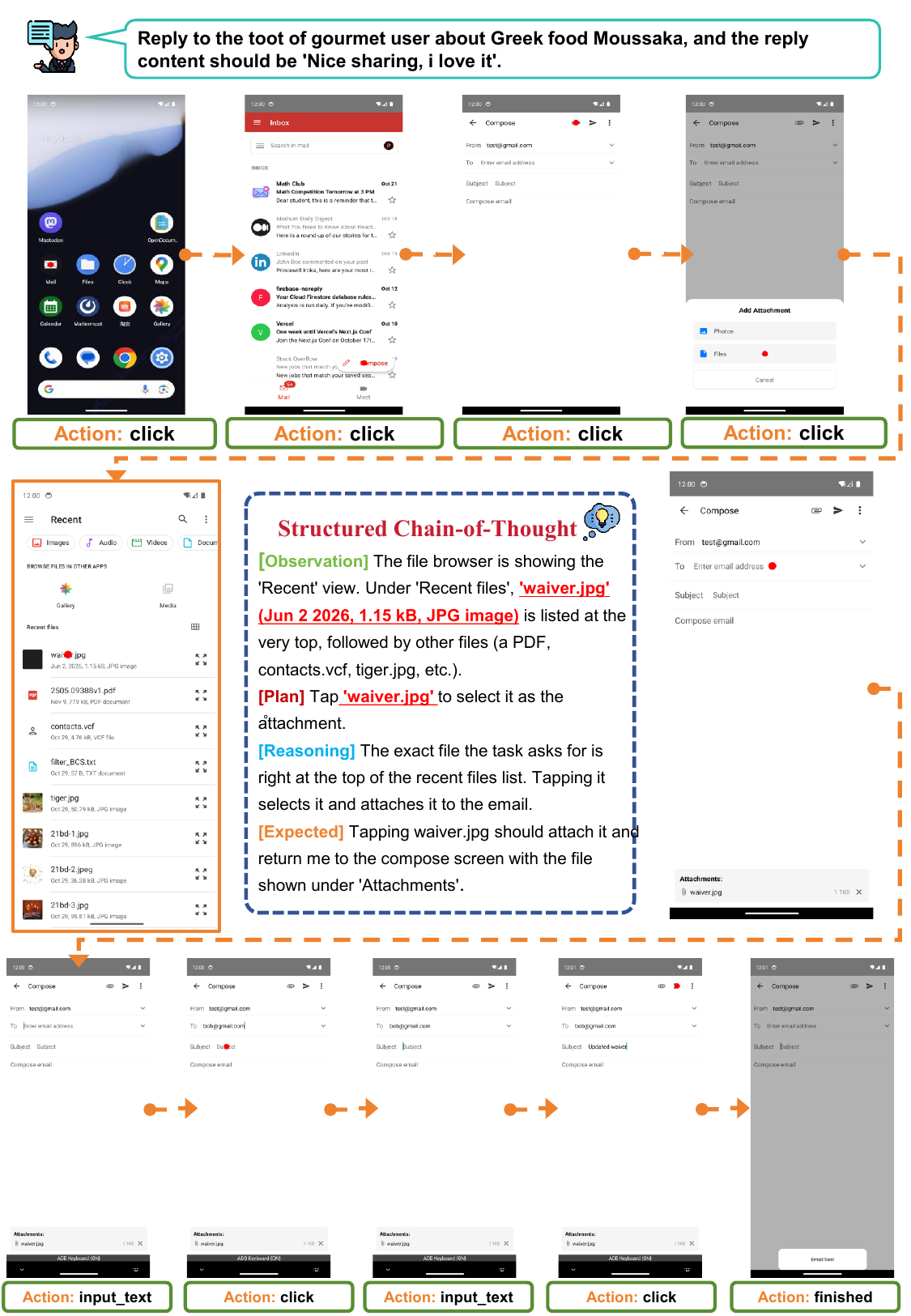}
  \caption{A representative mobile task executed by \METHODNAME. The model
  interprets the instruction, navigates across screens, and grounds its
  actions to the relevant UI elements.}
  \label{fig:mobilecase}
\end{figure}

Figure~\ref{fig:mobilecase} presents a mobile GUI task
executed by \METHODNAME. The trajectory illustrates three capabilities that
are important for mobile interaction: identifying the user's target from the
instruction, maintaining task progress while navigating between screens, and
grounding each action to an executable UI element. This example complements
the MobileWorld results by showing how the distilled student
applies mobile-specialist guidance in a complete interaction sequence.
Additional desktop cases are provided in
Appendix~\ref{sec:exp_case_study2}.

\FloatBarrier

\section{Conclusions}
\label{sec:conclusion}
This work studies specialist knowledge integration for a unified GUI agent
across heterogeneous desktop and mobile environments. We construct a
unified cross-platform data-collection harness and use it to build
\textbf{Uni-GUI}, which contains nearly 10K high-quality and executable
interaction trajectories. Building on Uni-GUI, we propose
\textbf{UI-MOPD}, the first framework to introduce multi-teacher on-policy
distillation into unified multi-platform GUI agent training. During
reinforcement learning, a shared student samples its own rollouts and routes
each rollout to the corresponding platform-specialized teacher. The routed
teacher acts as a behavioral anchor at student-visited states, integrating
complementary expertise without averaging distinct interaction conventions.
UI-MOPD achieves task success rates of 38.2\% on OSWorld and 12.0\% on
MobileWorld, attaining the strongest balanced performance among
parameter-matched 8B integration strategies while largely preserving
general GUI grounding. Our analyses further show that the two specialists
induce distinct capability shifts and provide useful corrections in
different platform states. These findings establish MOPD as an effective
approach to building a single cross-platform GUI agent.

\clearpage
\bibliographystyle{plainnat}
\bibliography{ref}

\clearpage
\phantomsection
\addcontentsline{toc}{section}{Appendix}
\addtocontents{toc}{\protect\setcounter{tocdepth}{-1}}
\beginappendix
\crefname{section}{Appendix}{Appendices}
\Crefname{section}{Appendix}{Appendices}
\appendix

\section{Dataset Construction and Composition}
\label{app:dataset_construction}

\paragraph{AndroidControl\textsuperscript{$\star$} Construction.}
To statically evaluate GUI understanding capability of \METHODNAME beyond
interactive task execution, we construct an extracted AndroidControl
subset, denoted as AndroidControl\textsuperscript{$\star$}. This subset
contains 4,260 step-level records from 781 Android trajectories. Each record
is stored in JSONL format and includes the trajectory identifier, step index,
high-level task goal, per-step instruction, normalized action, screenshot path,
and screenshot resolution. The referenced screenshots are provided as PNG files
and are linked directly from the corresponding step records.

AndroidControl\textsuperscript{$\star$} preserves the same normalized mobile
action space used in \BENCH, including click, scroll, input\_text, open\_app,
wait, navigate\_back, long\_press, and navigate\_home. For actions that can be
matched to a UI element, we also include grounding metadata such as target
bounding boxes, widget class, visible text or content description, resource id,
package name, ancestor information, and the number of matching UI nodes. Steps
without a directly groundable target, such as waiting or app-level operations,
leave the grounding field empty. This subset is mainly used to evaluate static
mobile GUI understanding, verify action-screen alignment, and illustrate the
format of mobile data after normalization.

\paragraph{\BENCH Composition.}

\BENCH is constructed from four groups of trajectories across desktop and
mobile platforms, as summarized in Table~\ref{tab:dataset_composition}. For
each platform, we combine trajectories collected by our Unified
Cross-Platform Data Collection Harness with cleaned open-source
trajectories. On the desktop side, the self-collected portion contains about
95K interaction steps from desktop GUI environments, and the public portion
contains about 13K cleaned steps from OpenCUA~\citep{Opencua}. On the mobile side, the
self-collected portion contains about 17K interaction steps from mobile GUI
environments, and the public portion contains about 35K cleaned steps from
OpenMobile~\citep{openmobile}. In total, \BENCH contains approximately 160K steps and 11.5K
trajectories.

We do not directly use raw public trajectories. Instead, OpenCUA and
OpenMobile are processed with the trajectory cleaning and post-processing
steps described in \cref{app:data_collection_harness}, including
action-space compatibility checking, trajectory filtering, and format
normalization.

\begin{table}[h]
\centering
\caption{Approximate composition of \BENCH. Counts are rounded for readability.}
\label{tab:dataset_composition}
\normalsize
\begin{adjustbox}{max width=\linewidth}
\begin{tabular}{llcc}
\toprule
\textbf{Platform} & \textbf{Source Type} & \textbf{Steps} & \textbf{Trajectories} \\
\midrule
\multirow{2}{*}{Desktop} & Self-collected & $\sim$95K & $\sim$7K \\
 & OpenCUA & $\sim$13K & $\sim$0.8K \\
\multirow{2}{*}{Mobile} & Self-collected & $\sim$17K & $\sim$1K \\
 & OpenMobile & $\sim$35K & $\sim$2.7K \\
\midrule
\multicolumn{2}{l}{Total} & $\sim$160K & $\sim$11.5K \\
\bottomrule
\end{tabular}
\end{adjustbox}
\end{table}

Including OpenCUA and OpenMobile broadens the coverage of GUI states,
applications, and task types beyond the self-collected trajectories. These
sources complement the desktop and mobile data collected by our harness,
increase platform and application diversity, and provide additional
supervision for cross-platform generalization after filtering and
normalization.

\section{Unified Cross-Platform Data Collection Harness}
\label{app:data_collection_harness}

Collecting high-quality GUI agent trajectories is challenging because the
validity of a trajectory depends on the textual instruction, the current
interface state, the executable action space, and the visual grounding of each
interaction. A task that appears reasonable at the language level may still be
unusable for training if the target UI element is not visible, the action
cannot be represented by the student policy, or the instruction is
inconsistent with the environment state. These issues become more pronounced
in a cross-platform setting, where desktop and mobile environments use
different observation formats, action primitives, and UI layouts. To reduce
such noise, the harness organizes data construction into four stages: query generation, 
trajectory collection, trajectory
cleaning, and post-processing.

\begin{figure}[H]
  \centering
  \includegraphics[width=\linewidth]{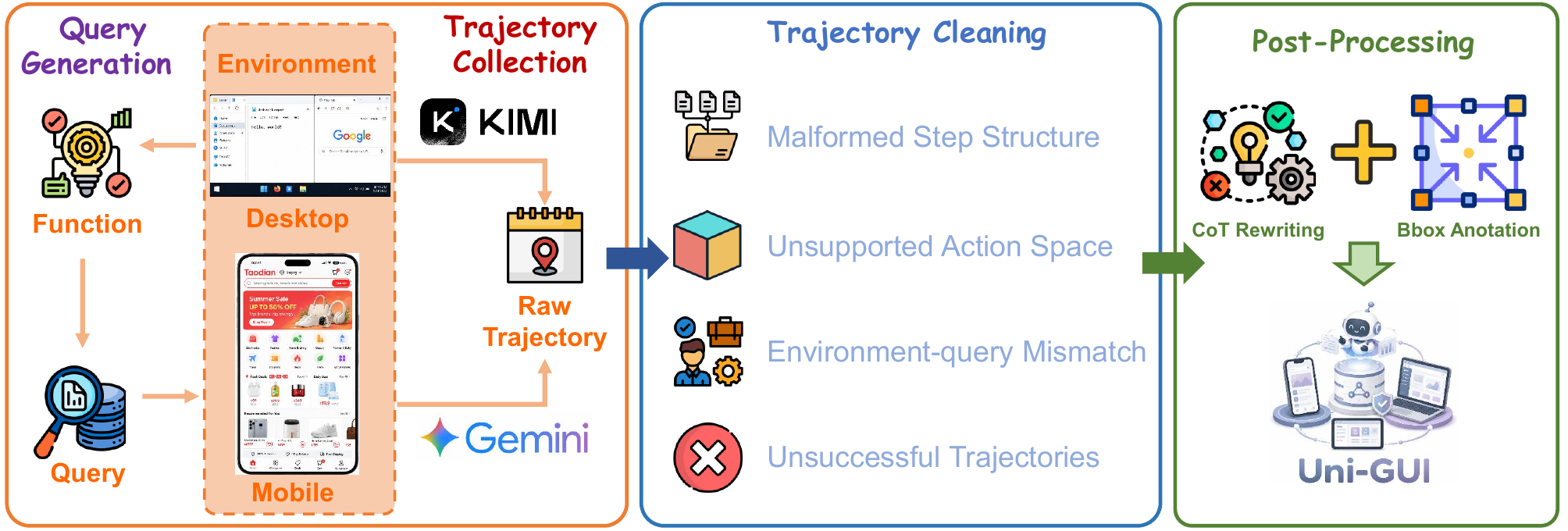}
  \caption{Overview of Unified Cross-Platform Data Collection Harness.}
  \label{fig:harness}
\end{figure}

\paragraph{Query Generation.}
We first construct user queries from executable functionalities in the target
environments. Instead of freely generating arbitrary instructions, we ask
strong teacher models to identify realistic functional points supported by the
corresponding desktop or mobile environment, and then synthesize natural user
queries from these functional points. In our implementation, desktop queries
are generated with the assistance of Kimi-K2.6, while mobile queries
are generated with the assistance of Gemini-3.1-Pro. This
environment-grounded query construction reduces the chance of collecting
trajectories for underspecified, infeasible, or state-mismatched tasks.

\paragraph{Query Generation.}
We first construct environment-grounded user queries from real, executable
functionalities in the target GUI environments. For the desktop domain,
Kimi-K2.6 is used to assist functional-point extraction from the
original initialized OSWorld~\citep{osworld} environments. For the mobile domain,
Gemini-3.1-Pro is used to assist functional-point extraction from
the original initialized MobileWorld~\citep{mobileworld} and AndroidWorld~\citep{androidworld} environments. Instead
of generating arbitrary instructions in free form, we ask the teacher models
to identify realistic and usable functionalities supported by these
environments. Based on the extracted functional points, we then synthesize
user queries that are both natural and executable in the corresponding
environments. This design reduces the likelihood of collecting trajectories
for tasks that are underspecified, infeasible, or mismatched with the
environment state.

\paragraph{Trajectory Collection.}
Given a generated query, the teacher model interacts with the target GUI
environment to produce an execution trajectory. Each trajectory records the
observations, actions, intermediate reasoning, and task states during the
rollout. Desktop and mobile trajectories are collected through the same
high-level interface, but the executed actions are kept in their native
platform forms before later normalization. This allows the harness to retain
platform-specific interaction patterns, such as desktop-oriented pointing and
window operations or mobile-oriented touch and navigation behaviors, while
keeping the collected data compatible with a unified training pipeline.

\paragraph{Trajectory Cleaning.}
Raw trajectories collected from GUI environments are noisy and cannot be
directly used for Qwen-VL student-model training. We therefore apply a multi-stage
cleaning pipeline. First, we remove trajectories with malformed step
structures, such as non-contiguous step indices or duplicated steps. Second,
we filter out trajectories whose actions cannot be mapped to the action space
of the student model. This ensures that every retained demonstration can be
faithfully imitated by the student. Third, we discard overly long trajectories
with more than 40 steps, since such trajectories are more likely to contain
inefficient exploration, accumulated errors, or ambiguous supervision signals.
Fourth, we remove trajectories whose queries are inconsistent with the original
environment or unsupported by the collected functional points.

Finally, we use Gemini-3.1-Pro as an automatic judge to check whether the trajectory
successfully completes the intended task, and only retain successful
trajectories. Specifically, we decompose each task query into an
ordered list of sub-tasks before judging. During trajectory inspection, the
judge walks through the execution steps, identifies which sub-task each step
corresponds to, and determines whether the corresponding sub-task has been
completed. A trajectory is retained only when all sub-tasks are judged as
completed. This sub-task-level adjudication avoids relying on a single
long-context judgment over the entire trajectory, reduces the effect of noisy
or redundant steps, and provides clearer attribution for failed executions.

\paragraph{Post-Processing.}
After cleaning, we convert the remaining trajectories into the training format
required by the Qwen-VL-based policy. We first normalize the intermediate
reasoning into a structured chain-of-thought format. This is necessary because
the raw reasoning traces produced by Kimi-K2.6 and
Gemini-3.1-Pro are often heterogeneous in structure, vary
substantially in length, and differ from the reasoning style of Qwen3-VL
models. Directly training on these unnormalized traces may disturb the
student model's original output distribution. Therefore, we rewrite the
reasoning traces into a consistent structure aligned with the expected
Qwen3-VL input-output format.

We also re-annotate grounding bounding boxes for actions that refer to visual
UI elements. These bounding boxes are used in the subsequent rule-based
evaluation stage to determine whether the current action is executed on the
correct target region. The final output of the harness is therefore a set of
successful, executable, action-compatible, reasoning-normalized, and visually
grounded GUI trajectories across both mobile and desktop platforms.

\section{Training and Implementation Details}

\subsection{Action Space and Trajectory Format}

The prompt templates in Appendix~\ref{app:prompt_templates} define the
platform specific tool interfaces used by Qwen3-VL based policies. We
summarize the corresponding action spaces in Table~\ref{tab:action_space}.
Desktop trajectories use the computer\_use interface with
mouse and keyboard actions, while mobile trajectories use the
mobile\_use interface with touchscreen actions. During
post processing, actions from different data sources are converted into these
platform specific tool call formats so that every retained trajectory can be consumed by the training pipeline.

\begin{table}[h]
\centering
\caption{Platform specific action spaces used in the trajectory format.}
\label{tab:action_space}
\normalsize
\setlength{\tabcolsep}{8pt}
\renewcommand{\arraystretch}{1.18}
\begin{tabular}{p{0.11\linewidth}p{0.15\linewidth}p{0.62\linewidth}}
\toprule
\textbf{Platform} & \textbf{Tool} & \textbf{Actions} \\
\midrule
\multirow{2}{*}{Desktop} &
\multirow{2}{*}{computer\_use} &
key, type, mouse\_move, left\_click, left\_click\_drag, right\_click \\
& & middle\_click, double\_click, triple\_click, scroll, wait, terminate \\
\addlinespace[5pt]
\multirow{2}{*}{Mobile} &
\multirow{2}{*}{mobile\_use} &
click, long\_press, swipe, type, answer \\
& & system\_button, wait, ask\_user, terminate \\
\bottomrule
\end{tabular}
\end{table}

Each trajectory is stored as an episode directory. A mobile episode
contains a normalized trajectory file task.json, a raw generation
record task\_raw.json, and screenshots indexed by step, such as
0.jpg and 1.jpg. The normalized
task.json file stores episode metadata, including the task
source, application name, application package, screen resolution, user query,
episode identifier, device type, and train/test split. Its data
field contains the step trajectory records. Each step records the step
index, query, normalized reasoning, tool call action plan, screen resolution,
grounding bounding boxes, screenshot path, and filtering or
review flags.

The raw file task\_raw.json is retained for traceability. It stores
the original prompt, raw model response, raw action, converted action, and
labeled screenshot reference for each step. This separation allows us to keep
the original generation evidence while training only on the cleaned and
normalized trajectory representation.

\subsection{Training Hyperparameters}

Table~\ref{tab:training_hyperparameters} summarizes the main training
configuration used in our experiments. The student policy is initialized from
Qwen3-VL-8B-Thinking, while the platform-specific teachers are
initialized from Qwen3-VL-32B-Thinking. Both teacher SFT and student
training are run for one epoch. Student training uses a GRPO-based DAPO
objective with multi-teacher on-policy distillation, where each prompt samples
8 rollouts and the OPD auxiliary KL loss is enabled with coefficient 0.01.
For visual inputs, training uses only the current screenshot, while inference
uses four historical screenshots, the current screenshot, and the full text
action history.
All training is conducted on 64 NVIDIA H100 GPUs organized as 8 nodes with
8 GPUs each, and asynchronous rollouts are served by SGLang.

\begingroup
\normalsize
\setlength{\tabcolsep}{4pt}
\renewcommand{\arraystretch}{0.96}
\begin{longtable}{lll}
\caption{Training hyperparameters. TP, PP, and DP denote tensor, pipeline, and data parallelism, respectively.}
\label{tab:training_hyperparameters}\\
\toprule
\textbf{Category} & \textbf{Hyperparameter} & \textbf{Value} \\
\midrule
\endfirsthead
\caption[]{Training hyperparameters.}\\
\toprule
\textbf{Category} & \textbf{Hyperparameter} & \textbf{Value} \\
\midrule
\endhead
\bottomrule
\endfoot
\rowcolor{black!8}
\multicolumn{3}{l}{\textit{Models and Training Epochs}} \\
Student model & Initialization & \texttt{Qwen3-VL-8B-Thinking} \\
Teacher model & Initialization & \texttt{Qwen3-VL-32B-Thinking} \\
Teacher SFT & Epochs & 1 \\
Student training & Epochs & 1 \\
\midrule
\rowcolor{black!8}
\multicolumn{3}{l}{\textit{Infrastructure}} \\
Cluster & Nodes / GPUs per node & 8 / 8 \\
Total GPUs & -- & 64 NVIDIA H100 GPUs \\
\midrule
\rowcolor{black!8}
\multicolumn{3}{l}{\textit{Parallelism}} \\
Student 8B & TP / PP / DP & 2 / 1 / 32 \\
Teacher 32B & TP / DP & 8 / 8 \\
Rollout & TP & 2 \\
\midrule
\rowcolor{black!8}
\multicolumn{3}{l}{\textit{Batch Size and Sequence Length}} \\
Training batch size & -- & 128 \\
Generation batch size & -- & 384 \\
Mini batch size & -- & 128 \\
Micro batch size per GPU & -- & 4 \\
Maximum prompt length & -- & 8192 \\
Maximum response length & -- & 512 \\
\midrule
\rowcolor{black!8}
\multicolumn{3}{l}{\textit{Visual Input}} \\
Desktop image resolution & Desktop & $1920 \times 1080$ \\
Mobile image resolution & Mobile & $1080 \times 2400$ \\
Training visual context & Screenshots & Current screenshot only \\
Inference visual context & Screenshots & 4 history screenshots + current screenshot \\
Inference text context & Action history & All previous text actions \\
Image pixel range & Min / max pixels & 3,136 / 13,107,200 \\
\midrule
\rowcolor{black!8}
\multicolumn{3}{l}{\textit{Optimization}} \\
Learning rate & -- & $1\times10^{-6}$ \\
Precision & -- & bfloat16 \\
Rollout samples per prompt & -- & 8 \\
Clip ratio & Low / high / C & 0.2 / 0.28 / 10.0 \\
Loss aggregation & -- & Token mean \\
\midrule
\rowcolor{black!8}
\multicolumn{3}{l}{\textit{OPD KL Loss}} \\
KL loss type & -- & k3 \\
KL loss coefficient & -- & 0.01 \\
\midrule
\rowcolor{black!8}
\multicolumn{3}{l}{\textit{Rollout Engine}} \\
Engine & -- & SGLang \\
Mode & -- & Async \\
Temperature / top-$p$ & -- & 1.0 / 1.0 \\
GPU memory utilization & -- & 0.60 \\
Maximum number of sequences & -- & 1024 \\
\end{longtable}
\endgroup

\section{Fine-Grained Static GUI and Grounding Results}
\label{app:fine_grained_static_grounding}

\begin{table}[H]
\centering
\caption{Fine-grained results on AndroidControl\textsuperscript{$\star$} and grounding benchmarks. AndroidControl\textsuperscript{$\star$} denotes the evaluated subset. Base denotes Qwen3-VL-8B-Thinking, and Model Merge corresponds to the TIES-merging checkpoint.}
\label{tab:appendix_fine_grained_static_grounding}
\normalsize
\setlength{\tabcolsep}{4pt}
\renewcommand{\arraystretch}{0.92}
\begin{adjustbox}{max width=\linewidth}
\begin{tabular}{lccc}
\toprule
\textbf{Benchmark / Metric} & \textbf{Base} & \textbf{Model Merge (TIES)} & \textbf{\METHODNAME} \\
\midrule
\rowcolor{black!8}
\multicolumn{4}{l}{\textit{AndroidControl\textsuperscript{$\star$}}} \\
Action Type Accuracy & 85.75\% & 81.62\% & 87.02\% \\
Grounding (target) & 88.04\% & 86.02\% & 88.33\% \\
Grounding (ancestor) & 89.59\% & 87.69\% & 89.98\% \\
Overall Accuracy & 78.73\% & 74.01\% & 80.05\% \\
\midrule
\rowcolor{black!8}
\multicolumn{4}{l}{\textit{ScreenSpot-Pro}} \\
Overall & 43.71\% & 37.13\% & 43.14\% \\
CAD & 24.90\% & 20.31\% & 22.61\% \\
Dev & 41.81\% & 32.78\% & 41.14\% \\
Creative & 41.06\% & 35.48\% & 41.64\% \\
Scientific & 50.39\% & 50.79\% & 52.36\% \\
Office & 64.78\% & 49.57\% & 63.48\% \\
OS & 42.86\% & 36.73\% & 40.31\% \\
\midrule
\rowcolor{black!8}
\multicolumn{4}{l}{\textit{ScreenSpotV2}} \\
Overall & 91.27\% & 88.60\% & 90.88\% \\
mobile & 93.41\% & 92.02\% & 91.62\% \\
desktop & 90.12\% & 88.02\% & 92.22\% \\
web & 89.70\% & 85.13\% & 89.02\% \\
\midrule
\rowcolor{black!8}
\multicolumn{4}{l}{\textit{OSWorld-G}} \\
Overall & 52.13\% & 47.16\% & 52.84\% \\
Text Matching & 31.58\% & 47.37\% & 42.11\% \\
Element Recognition & 59.70\% & 47.76\% & 57.46\% \\
Layout Understanding & 56.44\% & 53.78\% & 60.44\% \\
Fine-grained Manipulation & 51.52\% & 48.48\% & 50.76\% \\
Refusal & 24.07\% & 14.81\% & 18.52\% \\
\bottomrule
\end{tabular}
\end{adjustbox}
\end{table}

Table~\ref{tab:appendix_fine_grained_static_grounding} provides fine-grained
results for the static GUI and grounding evaluations. On
AndroidControl\textsuperscript{$\star$}, \METHODNAME improves all reported
metrics over the base model, including action type prediction, target
grounding, ancestor grounding, and overall accuracy. In contrast, Model Merge
consistently degrades these metrics, suggesting that static parameter merging
is less stable for preserving mobile GUI understanding.

For the grounding benchmarks, \METHODNAME largely preserves the base model's
performance while avoiding the larger degradation observed in Model Merge.
On ScreenSpot-Pro and ScreenSpotV2, \METHODNAME remains close to the base
model and improves several subcategories such as Creative, Scientific, and
desktop grounding. On OSWorld-G, \METHODNAME achieves the best overall score
and improves layout understanding, indicating that multi-teacher on-policy
distillation can retain fine-grained GUI grounding ability while improving
cross-platform interactive performance.

\section{Additional Case Studies}
\label{sec:exp_case_study2}

\begin{figure}[H]
  \centering
  \includegraphics[width=0.90\linewidth]{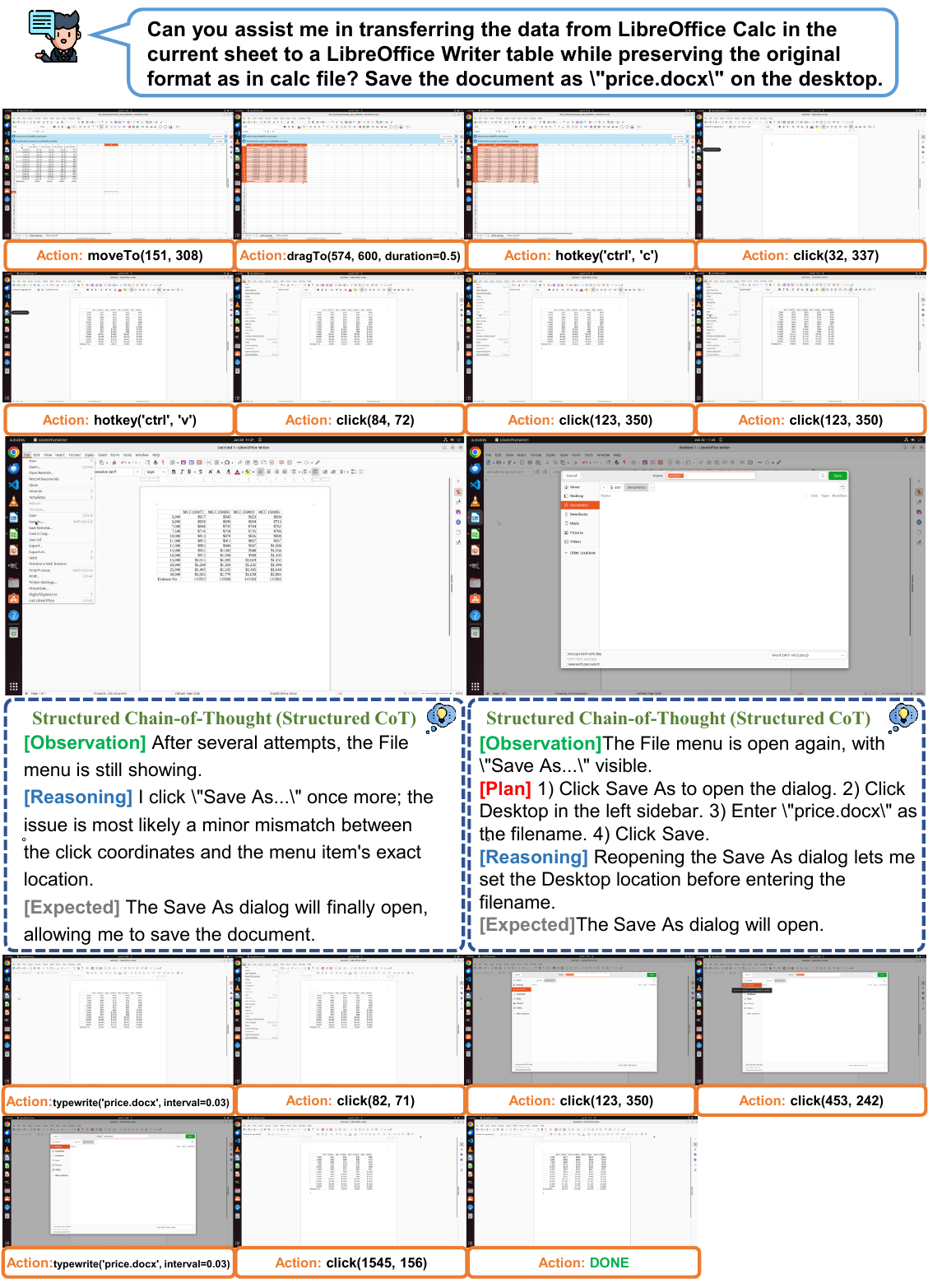}
  \caption{Desktop task execution example of \METHODNAME.}
  \label{fig:desktopcase}
\end{figure}

Figure~\ref{fig:desktopcase} shows an additional desktop GUI case study. The
example illustrates that \METHODNAME can follow a multi-step desktop
instruction, locate task-relevant regions in a dense interface, and execute
mouse-and-keyboard actions according to the current screen state. Compared
with mobile tasks, desktop tasks require handling larger layouts, window-level
operations, and more precise cursor-based grounding. Together with the mobile
case in Section~\ref{sec:exp_case_study}, this example qualitatively supports
the cross-platform capability of \METHODNAME.

\clearpage
\section{Prompt Templates}
\label{app:prompt_templates}

We provide four system prompts used in our data construction and policy
training pipeline. The first two prompts define the normalized desktop and
mobile tool interfaces for Qwen3-VL-based policies, which are used during
training and evaluation. The other two prompts are used for trajectory
collection: Kimi-K2.6 is used to collect desktop trajectories, while
Gemini-3.1-Pro is used to collect mobile trajectories before action
normalization and post-processing.

\begin{tcblisting}{
  enhanced,
  breakable,
  listing only,
  colback=black!2,
  colframe=xiaomievbg,
  title={Desktop System Prompt (Qwen3-VL)},
  fonttitle=\bfseries,
  arc=1mm,
  boxrule=0.6pt,
  left=1mm,
  right=1mm,
  top=1mm,
  bottom=1mm,
  listing options={
    basicstyle=\ttfamily\footnotesize,
    breaklines=true,
    columns=fullflexible,
    keepspaces=true
  }
}
# Tools

You may call one or more functions to assist with the user query.

You are provided with function signatures within <tools></tools> XML tags:
<tools>
{
  "type": "function",
  "function": {
    "name_for_human": "computer_use",
    "name": "computer_use",
    "description": "Use a mouse and keyboard to interact with a computer, and take screenshots. This is an interface to a desktop GUI. You do not have access to a terminal or applications menu. You must click on desktop icons to start applications. Some applications may take time to start or process actions, so you may need to wait and take successive screenshots. The screen resolution is 1000x1000. Consult screenshots before moving the cursor, and click UI elements with the cursor tip near the center.",
    "parameters": {
      "properties": {
        "action": {
          "enum": ["key", "type", "mouse_move", "left_click", "left_click_drag", "right_click", "middle_click", "double_click", "triple_click", "scroll", "wait", "terminate"]
        },
        "keys": {"description": "Required only by action=key."},
        "text": {"description": "Required only by action=type."},
        "coordinate": {"description": "The x,y coordinates for mouse actions."},
        "pixels": {"description": "The amount of scrolling."},
        "time": {"description": "The seconds to wait."},
        "status": {"enum": ["success", "failure"]}
      },
      "required": ["action"]
    }
  }
}
</tools>

For each function call, return a JSON object with function name and arguments within <tool_call></tool_call> XML tags:
<tool_call>
{"name": <function-name>, "arguments": <args-json-object>}
</tool_call>

# Response format

Response format for every step:
1) Action: a short imperative describing what to do in the UI.
2) A single <tool_call>...</tool_call> block containing only the JSON:
   {"name": <function-name>, "arguments": <args-json-object>}.

Rules:
- Output exactly in the order: Action, <tool_call>.
- Be brief: one sentence for Action.
- Do not output anything else outside those parts.
- If finishing, use action=terminate in the tool call.
\end{tcblisting}

\clearpage
\begin{tcblisting}{
  enhanced,
  breakable,
  listing only,
  colback=black!2,
  colframe=xiaomievbg,
  title={Mobile System Prompt (Qwen3-VL)},
  fonttitle=\bfseries,
  arc=1mm,
  boxrule=0.6pt,
  left=1mm,
  right=1mm,
  top=1mm,
  bottom=1mm,
  listing options={
    basicstyle=\ttfamily\footnotesize,
    breaklines=true,
    columns=fullflexible,
    keepspaces=true
  }
}
# Tools

You may call one or more functions to assist with the user query.

You are provided with function signatures within <tools></tools> XML tags:
<tools>
{
  "type": "function",
  "function": {
    "name": "mobile_use",
    "description": "Use a touchscreen to interact with a mobile device, and take screenshots. This is an interface to a mobile device with touchscreen. You can perform actions like clicking, typing, swiping, etc. Some applications may take time to start or process actions, so you may need to wait and take successive screenshots. The screen resolution is 999x999. Click buttons, links, and icons near the center of the element.",
    "parameters": {
      "properties": {
        "action": {
          "enum": ["click", "long_press", "swipe", "type", "answer", "system_button", "wait", "ask_user", "terminate"]
        },
        "coordinate": {"description": "Required only by action=click, action=long_press, and action=swipe."},
        "coordinate2": {"description": "Required only by action=swipe."},
        "text": {"description": "Required only by action=type, action=ask_user, and action=answer."},
        "time": {"description": "Required only by action=long_press and action=wait."},
        "button": {"enum": ["Back", "Home", "Menu", "Enter"]},
        "status": {"enum": ["success", "failure"]}
      },
      "required": ["action"]
    }
  }
}
</tools>

For each function call, return a JSON object with function name and arguments within <tool_call></tool_call> XML tags:
<tool_call>
{"name": <function-name>, "arguments": <args-json-object>}
</tool_call>

# Response format

Response format for every step:
1) Thought: one concise sentence explaining the next move.
2) Action: a short imperative describing what to do.
3) A single <tool_call>...</tool_call> block containing only the JSON:
   {"name": <function-name>, "arguments": <args-json-object>}.

Rules:
- Output exactly in the order: Thought, Action, <tool_call>.
- Be brief: one sentence for Thought, one sentence for Action.
- Do not output anything else outside those three parts.
- If finishing, use mobile_use with action=terminate in the tool call.
\end{tcblisting}

\clearpage
\begin{tcblisting}{
  enhanced,
  breakable,
  listing only,
  colback=black!2,
  colframe=black!85,
  colbacktitle=black!85,
  coltitle=white,
  title={Desktop Trajectory Collection Prompt (Kimi-K2.6)},
  fonttitle=\bfseries,
  arc=1mm,
  boxrule=0.6pt,
  left=1mm,
  right=1mm,
  top=1mm,
  bottom=1mm,
  listing options={
    basicstyle=\ttfamily\footnotesize,
    breaklines=true,
    columns=fullflexible,
    keepspaces=true
  }
}
System Prompt

You are a GUI agent. You are given an instruction, a screenshot of the screen,
and your previous interactions with the computer. You need to perform a series
of actions to complete the task. The password of the computer is {password}.

For each step, provide your response in this format:
{thought}
## Action:
{action}
## Code:
{code}

In the code section, the code should be either pyautogui code or one of the
following functions wrapped in the code block:

- {
    "name": "computer.wait",
    "description": "Make the computer wait for 20 seconds for installation,
    running code, etc.",
    "parameters": {
      "type": "object",
      "properties": {},
      "required": []
    }
  }

- {
    "name": "computer.terminate",
    "description": "Terminate the current task and report its completion status",
    "parameters": {
      "type": "object",
      "properties": {
        "status": {
          "type": "string",
          "enum": ["success", "failure"],
          "description": "The status of the task"
        },
        "answer": {
          "type": "string",
          "description": "The answer of the task"
        }
      },
      "required": ["status"]
    }
  }
\end{tcblisting}

\clearpage
\begin{tcblisting}{
  enhanced,
  breakable,
  listing only,
  colback=blue!2!violet!3,
  colframe=blue!45!violet!85!black,
  colbacktitle=blue!45!violet!85!black,
  coltitle=white,
  title={Mobile Trajectory Collection Prompt (Gemini-3.1-Pro)},
  fonttitle=\bfseries,
  arc=1mm,
  boxrule=0.6pt,
  left=1mm,
  right=1mm,
  top=1mm,
  bottom=1mm,
  listing options={
    basicstyle=\ttfamily\footnotesize,
    breaklines=true,
    columns=fullflexible,
    keepspaces=true
  }
}
# Role: Android Phone Operator AI

You are an AI that controls an Android phone to complete user requests.
Your responsibilities:
- Answer questions by retrieving information from the phone.
- Perform tasks by executing precise actions.

# Action Framework

Respond with exact JSON format for one of these actions:
| Action | Description | JSON Format Example |
| click | Tap visible element | {"action_type": "click", "coordinate": [x, y]} |
| double_tap | Double tap visible element | {"action_type": "double_tap", "coordinate": [x, y]} |
| long_press | Long press visible element | {"action_type": "long_press", "coordinate": [x, y]} |
| drag | Drag from one visible element to another | {"action_type": "drag", "start_coordinate": [x1, y1], "end_coordinate": [x2, y2]} |
| input_text | Type into field | {"action_type": "input_text", "text": "Hello"} |
| answer | Respond to user | {"action_type": "answer", "text": "It's 25 degrees today."} |
| navigate_home | Return to home screen | {"action_type": "navigate_home"} |
| navigate_back | Navigate back | {"action_type": "navigate_back"} |
| scroll | Scroll direction | {"action_type": "scroll", "direction": "down"} |
| status | Mark task as complete or infeasible | {"action_type": "status", "goal_status": "complete"} |
| wait | Wait for screen to update | {"action_type": "wait"} |
| ask_user | Ask user for information | {"action_type": "ask_user", "text": "what is the exact requirements do you need?"} |
| keyboard_enter | Press enter key | {"action_type": "keyboard_enter"} |
Note:
- The coordinate is the center of the element to be clicked, long pressed, or dragged.
- x and y are screen coordinates, with origin at the top left corner.
- x and y are normalized to [0, 1000].
# Execution Principles
1. Communication Rule:
- Always use the answer action to reply to users.
- Follow the user instruction strictly, e.g., only return a single number,
  only return True or False, or only return items separated by comma.
- Never use answer to indicate waiting or loading; use wait instead.
- The answer action terminates the task immediately.
2. Efficiency First:
- Choose the simplest path to complete tasks.
- If an action fails twice, try alternatives, e.g., long_press instead of click.
3. Smart Navigation:
- Gather information when needed.
- For scrolling, scroll direction is inverse to swipe direction.
- If scroll fails, try the opposite direction.
4. Text Operations:
- First click the input box to activate it before typing.
- For text manipulation, long press to select, use selection bar options,
  and delete by selecting then cutting.
5. Ask User:
- If there is not enough information to complete the task, use ask_user.
# Decision Process
1. Analyze goal, history, and current screen.
2. Determine if the task is already complete, and use status if true.
3. If not, choose the most appropriate action.
4. Output in the exact format below. The action must be a valid JSON string.
5. Only one tool call is allowed in one action.
# Expected Output Format

Thought: [Analysis including reference to key steps or points when applicable]
Action: [Single JSON action]

# Output Format Example
Thought: I need to type the weather question into the search box.
Action: {"action_type": "input_text", "text": "What is weather like in San Francisco today?"}
\end{tcblisting}

\addtocontents{toc}{\protect\setcounter{tocdepth}{2}}

\end{document}